\documentclass[runningheads]{llncs}

\usepackage{eccv}

\usepackage{eccvabbrv}

\usepackage{graphicx}

\usepackage{booktabs}
\usepackage{algorithm}
\usepackage{amsmath}
\usepackage{algpseudocode}
\usepackage{xcolor}
\usepackage{multirow}
\usepackage{tabularx}
\newcolumntype{Y}{>{\centering\arraybackslash}X}
\usepackage[accsupp]{axessibility}  % Improves PDF readability for those with disabilities.

\usepackage[textsize=tiny]{todonotes}
\usepackage{ifthen}
\newboolean{comments}
\setboolean{comments}{false} % Set to false to remove comments

\usepackage{hyperref}

\usepackage{orcidlink}

\begin{document}

% ---------------------------------------------------------------
% TODO REVIEW: Replace with your title
\title{SAGE: Sink-Aware Grounded Decoding for Multimodal Hallucination Mitigation} 

% TODO REVIEW: If the paper title is too long for the running head, you can set
% an abbreviated paper title here. If not, comment out.
\titlerunning{SAGE}

% TODO FINAL: Replace with your author list. 
% Include the authors' OCRID for the camera-ready version, if at all possible.
% \author{First Author\inst{1}\orcidlink{0000-1111-2222-3333} \and
% Second Author\inst{2,3}\orcidlink{1111-2222-3333-4444} \and
% Third Author\inst{3}\orcidlink{2222--3333-4444-5555}}

% % TODO FINAL: Replace with an abbreviated list of authors.
% \authorrunning{Tripti Shukla et al.}
% % First names are abbreviated in the running head.
% % If there are more than two authors, 'et al.' is used.

% % TODO FINAL: Replace with your institution list.
% \institute{Princeton University, Princeton NJ 08544, USA \and
% Springer Heidelberg, Tiergartenstr.~17, 69121 Heidelberg, Germany
% \email{lncs@springer.com}\\
% \url{http://www.springer.com/gp/computer-science/lncs} \and
% ABC Institute, Rupert-Karls-University Heidelberg, Heidelberg, Germany\\
% \email{\{abc,lncs\}@uni-heidelberg.de}}
\author{Tripti Shukla \and Zsolt Kira}

\authorrunning{T. Shukla and Z. Kira}

\institute{
Georgia Institute of Technology\\
\email{\{tshukla9,zkira\}@gatech.edu}
}
\maketitle
% \zk{IMPORTANT: Comment out the todonotes section and make sure you put back line numbers (line 214 in eccv.sty)!}

\begin{abstract}
Large vision-language models (VLMs) frequently suffer from hallucinations, generating content that is inconsistent with visual inputs. Existing methods typically address this problem through post-hoc filtering, additional training objectives, or external verification, but they do not intervene during the decoding process when hallucinations arise. In this work, we introduce \textit{\textbf{SAGE}, a \textbf{S}ink-\textbf{A}ware \textbf{G}rounded D\textbf{e}coding framework} that mitigates hallucinations by dynamically modulating self-attention during generation. Hallucinations are strongly correlated with \emph{attention sink tokens} - punctuation or function tokens that accumulate disproportionate attention despite carrying limited semantic content. SAGE leverages these tokens as anchors to monitor grounding reliability in real time. At each sink trigger, the method extracts semantic concepts from the generated sequence, estimates their visual grounding using both self-attention maps and gradient-based attribution, and measures their spatial agreement. Based on this signal, self-attention distributions are adaptively sharpened or broadened to reinforce grounded regions or suppress unreliable ones. Extensive experiments across diverse hallucination benchmarks demonstrate that SAGE consistently outperforms existing decoding strategies, achieving substantial reductions in hallucination while preserving descriptive coverage, without requiring model retraining or architectural modifications. Our method achieves an average relative improvement of 10.65\% on MSCOCO and 7.19\% on AMBER across diverse VLM architectures, demonstrating consistent gains in hallucination mitigation.

\end{abstract}

\vspace{-10mm}
\section{Introduction}
\label{sec:intro}

\begin{figure}[!t]
 \centering
 \captionsetup{type=figure}
 \includegraphics[width=\textwidth]
 {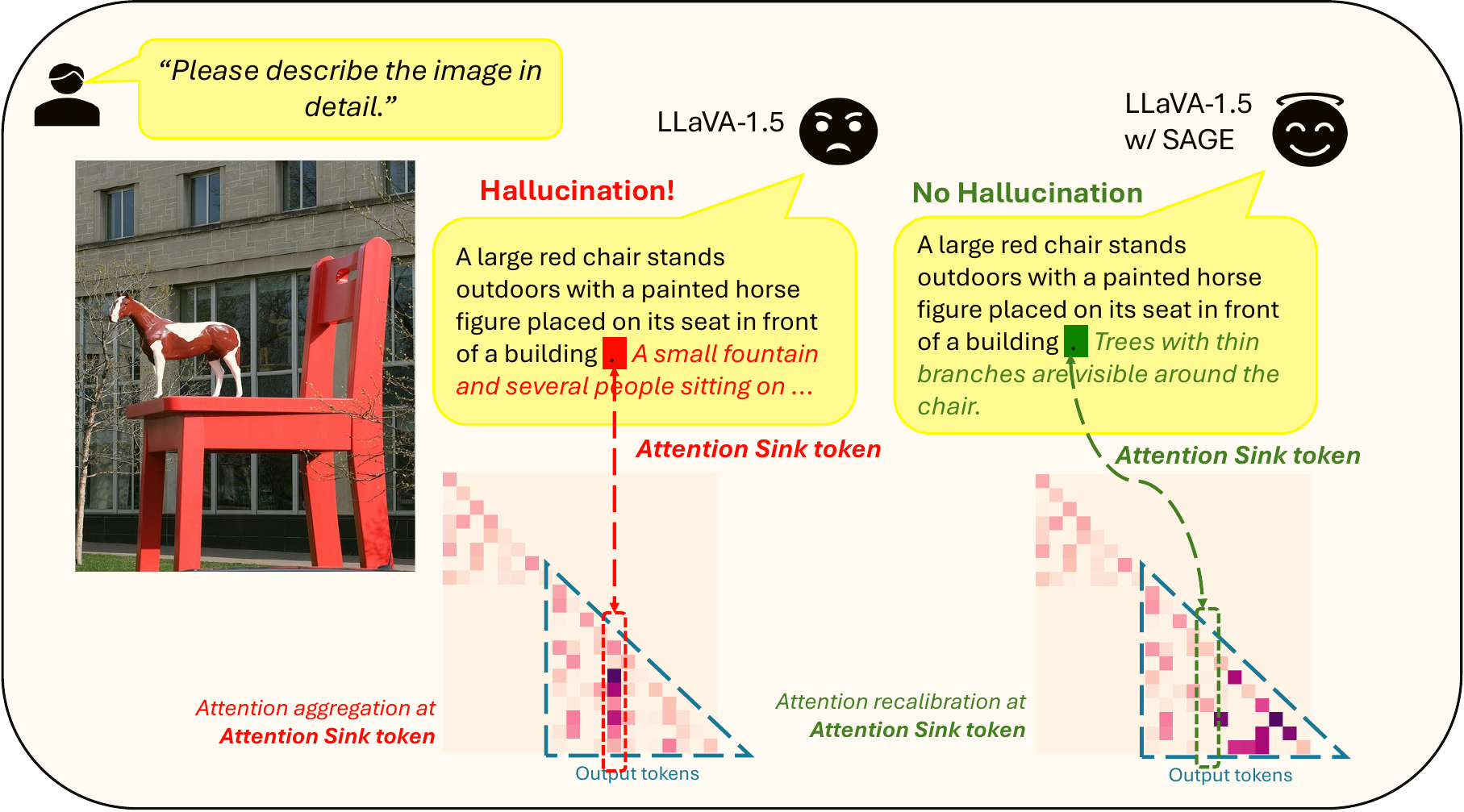}
 %\vspace{-20pt}
 \caption{We propose \textbf{SAGE}, a \emph{sink-aware grounded decoding strategy} for mitigating hallucinations in VLMs. SAGE detects \textbf{sink tokens} during generation and dynamically recalibrates attention towards image tokens to produce visually grounded descriptions.}
 \label{fig:teaser}
 \vspace{-6mm}
\end{figure}

Large vision-language models (VLMs) \cite{liu2023visualinstructiontuning, li2023blip2bootstrappinglanguageimagepretraining, wang2024qwen2vlenhancingvisionlanguagemodels, dai2023instructblipgeneralpurposevisionlanguagemodels,zhu2023minigpt4enhancingvisionlanguageunderstanding,chen2023shikraunleashingmultimodalllms} achieve strong multimodal understanding and generation, yet frequently suffer from multimodal hallucination \cite{liu2024surveyhallucinationlargevisionlanguage}, producing content unsupported by the image. This limits their reliability in real-world deployment. 
Existing mitigation methods either require retraining with additional supervision \cite{li2024finetuningmultimodalllmsfollow, liu2024mitigatinghallucinationlargemultimodal, park2024mitigatingdialoguehallucinationlarge} or applying training-free decoding heuristics \cite{liu2024payingattentionimagetrainingfree, li2025mitigatinghallucinationlargevision, wu2025mitigatinghallucinationsmultimodalspatial, fang2025groundinglanguagevisionconditional}. The former increases cost and complexity, while the latter typically operates at the output level without exploiting internal grounding signals. 

Recent works have identified \textit{attention sinks} - tokens that accumulate disproportionately high attention despite weak semantic contribution \cite{xiao2024efficientstreaminglanguagemodels, sun2024massiveactivationslargelanguage, kang2025toldvisualattentionsink} as a structural phenomenon in transformers. In multimodal models, these tokens often emerge during decoding and can dominate the attention distribution, effectively diverting the model’s focus away from relevant visual evidence. While several works have connected attention sinks to hallucination behavior in VLMs \cite{huang2024operaalleviatinghallucinationmultimodal, zhang2024seeingclearlylayertwo, liu2024payingattentionimagetrainingfree, tang2025seeingfarclearlymitigating}, they primarily treat them as artifacts to suppress, rather than potential signals that reveal when grounding failures are likely to occur. Also, most approaches rely solely on attention weights and do not incorporate complementary grounding cues from the vision encoder, limiting their ability to reliably detect misalignment between generated text and image content.

As depicted in Figure~\ref{fig:teaser}, we propose \emph{SAGE: Sink-Aware Grounded Decoding}, a training-free inference strategy that reinterprets sink tokens as grounding checkpoints during autoregressive generation. Our design is motivated by an empirical analysis on MSCOCO~\cite{lin2015microsoftcococommonobjects} and HalluBench~\cite{zhao2024hallucinationsenhancinglvlmshallucinationaware}, where we observe that \textbf{88.4\% of hallucination errors occur within five decoding steps after the generation of a sink token}. This strong temporal correlation suggests that sink tokens frequently mark transition boundaries where the model shifts from describing one concept to initiating another - points at which grounding failures are more likely to occur.

At each sink token, SAGE evaluates the visual grounding of the preceding segment using (i) multimodal self-attention over image tokens and (ii) gradient-based visual attribution. Their spatial agreement yields a real-time reliability signal: high agreement reinforces attention over grounded regions, while low agreement redistributes attention toward alternative visual evidence. By selectively modulating self-attention without retraining or architectural changes, SAGE effectively reduces hallucination while preserving descriptive quality.
% \zkn{This is an extremely short intro. Usually there is much more narrative here (at least a page). }

\textbf{Contributions:} (1) We propose SAGE, a training-free decoding framework that leverages attention sink tokens as triggers for real-time visual grounding verification during generation. (2) We introduce a reliability-driven attention modulation mechanism based on the agreement between self-attention and gradient-based attribution to improve grounding. (3) Experiments on five LVLMs across multiple benchmarks show that SAGE consistently outperforms existing hallucination mitigation methods.

% We introduce sink-aware grounded decoding, a test-time method that mitigates multimodal hallucinations by modulating attention based on concept-level visual grounding signals. (2) We analyze transformer layers and gradient-based visual attribution, and show that our approach consistently improves grounding and reduces hallucinations across vision–language benchmarks.              
\vspace{-3mm}
\section{Related Works}
% \vspace{-4mm}
\noindent\textbf{VLM Hallucination Mitigation.} A central cause of hallucination in VLMs is over-reliance on linguistic priors rather than visual evidence \cite{wang2025mllmseedynamiccorrection, zhang2025debiasingmultimodallargelanguage}. Training-time approaches mitigate this by introducing specialized modules \cite{zhao2024lookingtextreducinglanguage} or curating datasets and augmentations that emphasize visual features \cite{pi2024strengtheningmultimodallargelanguage, chen2025perturbollavareducingmultimodalhallucinations}; for example, HACL~\cite{jiang2024hallucinationaugmentedcontrastivelearning} employs contrastive learning over multimodal representations to distinguish hallucinated responses from factual ones. However, such methods typically incur substantial computational and engineering overhead. In contrast, inference-time methods are more flexible and can be broadly grouped into three families. Contrastive decoding \cite{leng2023mitigatingobjecthallucinationslarge, wang2025mintmitigatinghallucinationslarge, zhu2024ibdalleviatinghallucinationslarge} compares outputs under original and perturbed visual inputs to downweight prior-driven predictions, but requires additional forward passes for counterpart generations. Visual input modification emphasizes on the salient regions by altering the image (e.g., blurring, zooming, or cropping guided by intermediate signals) \cite{yu2024attentionpromptingimagelarge, mao2025magnifyingglassadaptiveperception, zhang2025mllmsknowlooktrainingfree}, which adds computational overhead and may struggle when evidence is spatially dispersed. Attention steering directly amplifies attention to visual tokens through fixed or proportional scaling \cite{yin2025clearsightvisualsignalenhancement, chen2025spatialreasoninghardvlms, zhang2024seeingclearlylayertwo, liu2024payingattentionimagetrainingfree}. Our method also operates purely at inference time, but instead strengthens visual contributions and query–image alignment within intermediate layers without modifying the input.

\noindent \textbf{Attention Sink.} Attention sink refers to the phenomenon where certain tokens accumulate disproportionately high attention despite having limited semantic relevance. This behavior has been observed across LLMs~\cite{xiao2024efficientstreaminglanguagemodels, yu2024unveilingharnessinghiddenattention, ferrando2024informationflowroutesautomatically}, vision transformers~\cite{darcet2024visiontransformersneedregisters}, and VLMs~\cite{kang2025toldvisualattentionsink, zhu2025mitigatingobjecthallucinationslarge}. Existing approaches primarily mitigate this issue by recalibrating or redistributing attention during inference. For instance, OPERA~\cite{huang2024operaalleviatinghallucinationmultimodal} penalizes columnar attention patterns during beam search to reduce reliance on summary tokens, while DOPRA~\cite{Wei_2024} refines this strategy through weighted overlay penalties and layer-wise redistribution. EAH~\cite{zhang2024seeingclearlylayertwo} mitigates vision sink effects by redistributing attention in shallow layers, whereas GIFT~\cite{qi2025capturinggazeshiftsguidance} leverages visual saliency maps to amplify attention toward visually relevant regions.

\vspace{-2mm}
\section{Preliminaries}

\noindent\textbf{Large Vision-Language Models.}
Large vision-language models (VLMs) such as LLaVA~\cite{liu2024improvedbaselinesvisualinstruction} consist of a visual encoder (e.g., CLIP~\cite{radford2021learningtransferablevisualmodels}), a pretrained large language model (LLM), and a projection module (e.g., Q-Former~\cite{li2023blip2bootstrappinglanguageimagepretraining}, MLP~\cite{chen2020simpleframeworkcontrastivelearning}) that maps visual features into the language embedding space. Given an image $I$, the encoder produces vision tokens $x_V$, which are combined with textual tokens $x_T$ and processed by the LLM for autoregressive generation:
\[
y_t = \arg\max p_\theta(y_t \mid y_{<t}, x_V, x_T),
\]
where $y_{<t}$ denotes previously generated tokens.

\noindent\textbf{Transformer Attention.}
The LLM is a transformer with stacked multi-head self-attention layers. For head $i$ in layer $l$, attention is computed as
\[
A_{l,i}(X_l) = \mathrm{softmax}\!\left(\frac{QK^\top}{\sqrt{d_k}}\right)V,
\]
where $Q,K,V$ are linear projections of the input tokens. During autoregressive generation, attention operates jointly over visual tokens $x_V$ and previously generated text $y_{<t}$. We denote the attention output corresponding to token $y_t$ at layer $l$ and head $i$ as $A_{l,i}(y_t \mid y_{<t}, x_V, x_T)$, which captures the alignment between generated tokens and visual context.

\section{Attention Sink}

% \zkn{Given ECCV's format, we will probably want to reduce subsections for space and use textbf instead}In autoregressive transformer decoding, attention sinks are tokens that accumulate disproportionately high incoming attention from subsequent tokens across multiple heads and layers \cite{xiao2024efficientstreaminglanguagemodels}\zkn{cite}. These tokens act as aggregation hubs where contextual information is consolidated before the model proceeds to generate new semantic units.\zkn{Tie this to hallucinations, i.e. intuition for why these tokens might be related to hallucinations and how your method will leverage that}

Transformer-based models rely on self-attention to distribute contextual information across tokens during autoregressive decoding \cite{vaswani2023attentionneed}. However, empirical analysis show that attention is often unevenly distributed: a small subset of tokens consistently accumulates disproportionately high incoming attention across layers and heads. This behavior, referred to as the \emph{attention sink} phenomenon, was first identified in large language models \cite{xiao2024efficientstreaminglanguagemodels}, where initial tokens attract excessive attention despite limited semantic contribution.

A key reason for this behavior lies in the Softmax normalization within self-attention. Since attention weights must sum-to-one, and initial tokens remain visible at every decoding step, they repeatedly absorb redundant attention. Over time, this leads to persistent columnar attention patterns, where subsequent tokens route a significant portion of their attention to these sink tokens, effectively turning them into aggregation hubs.

In multimodal large vision-language models (LVLMs), this imbalance becomes particularly problematic because textual and visual tokens compete within the same attention space and excessive routing toward sink tokens reduces attention allocated to image tokens. As a result, visual evidence is underutilized, and the model increasingly relies on autoregressive textual priors. This imbalance can dominate token prediction and ultimately manifest as hallucination \cite{huang2024operaalleviatinghallucinationmultimodal}. These observations motivate the need for decoding strategies that explicitly identify and regulate attention sinks to preserve visual grounding and mitigate hallucinations.

% \zkn{This is great! And should go first. Talk about sink tokens (from literature), hypothesize that hallucinations can occur at them because X, Y, Z, and then say ``To verify this, we...'' and descripe these experiments. }
\noindent\textbf{Sink Token Analysis.} \label{sink_tok_analysis} Our focus on attention sink tokens is motivated by an empirical analysis on MSCOCO~\cite{lin2015microsoftcococommonobjects} and HalluBench~\cite{zhao2024hallucinationsenhancinglvlmshallucinationaware}. We observe that 88.4\% of hallucination errors occur within five decoding steps after the generation of a sink token. This strong temporal correlation indicates that sink tokens frequently mark transition boundaries where the model shifts from describing one concept to initiating another - points at which grounding failures are more likely to occur.

At these transition points, the model aggregates substantial contextual attention from preceding tokens while contributing minimal new semantic content. Prior work has shown that punctuation tokens often behave as \emph{attention sinks}, accumulating disproportionately high attention despite weak semantic meaning~\cite{barbero2025llmsattendtoken, chauhan2025punctuationpredicateslanguagemodels, guo2024deepseekcoderlargelanguagemodel}. Such tokens - typically punctuation marks (e.g., periods, commas, colons, semicolons, dashes, ellipses) and lightweight conjunctions (e.g., ``and'', ``or'') exhibit persistent columnar attention patterns and act as aggregation hubs during decoding. Our empirical analysis in multimodal generation reveals a similar phenomenon, where these structurally lightweight tokens consistently attract high attention and frequently precede hallucination errors.

Based on this observation, we treat these sink tokens as hallucination prediction anchors during decoding. Whenever a sink token is generated, it triggers a targeted grounding verification step in our framework. By intervening at these natural structural boundaries, our method proactively detects and mitigates hallucination before it propagates to subsequent tokens.
% \zkn{Great description/motivation!}

\begin{figure*}[!t]
 \centering
 \captionsetup{type=figure}
 \includegraphics[width=\textwidth]
 {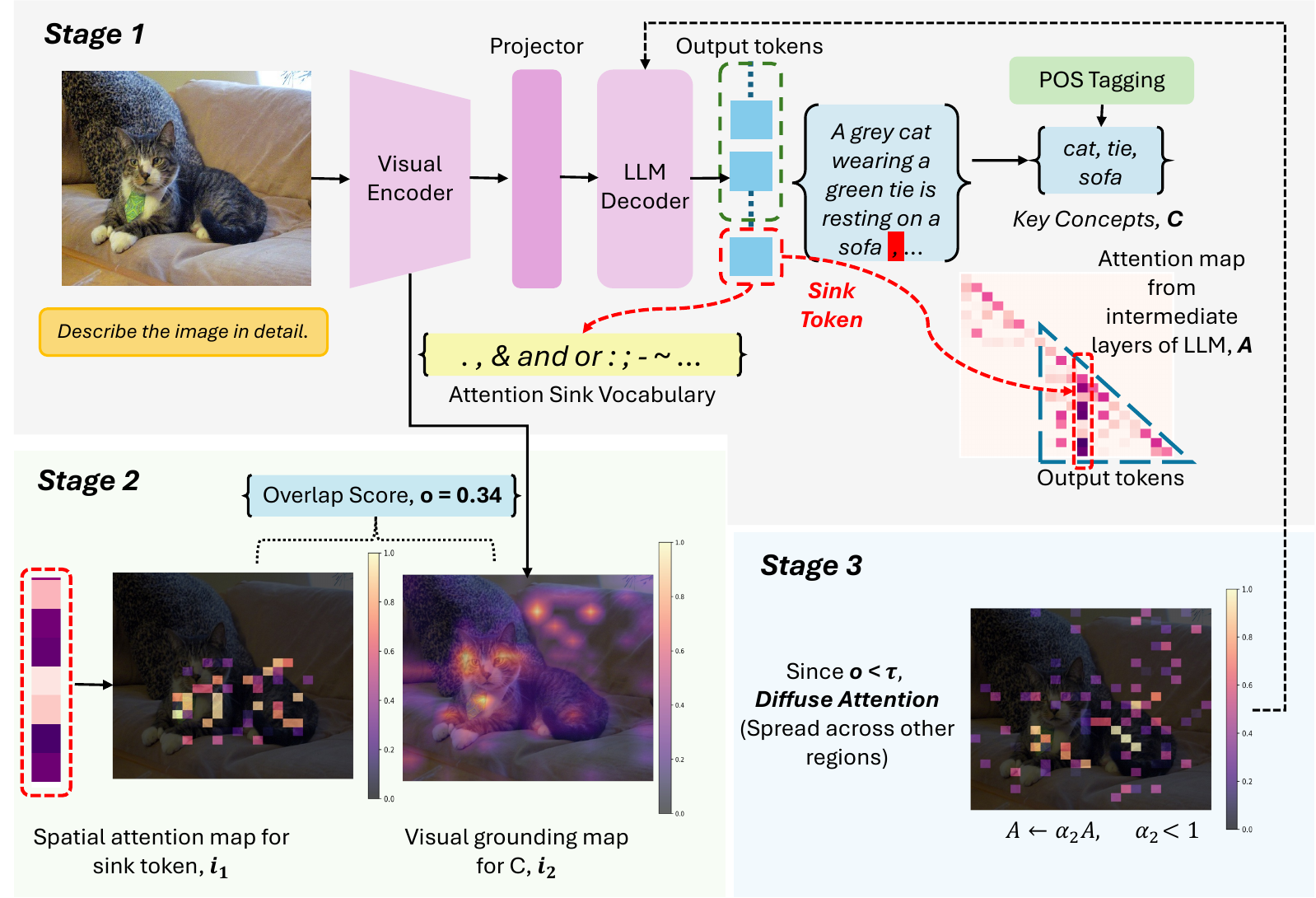}
 %\vspace{-20pt}
 \caption{
Overview of our inference-time decoding strategy, \textbf{SAGE}. 
\textbf{Stage 1:} SAGE identifies attention sink tokens and extracts key semantic concepts $C$ from the partially generated output. 
\textbf{Stage 2:} We project the sink-token attention onto the image to obtain an attention-based grounding map $i_1$, and compute concept-level visual grounding $i_2$ using Grad-CAM from the final layer of the vision encoder. An overlap score $o$ is then calculated between the two maps. 
\textbf{Stage 3:} Based on a threshold $\tau$, SAGE adaptively modulates attention: when $o < \tau$, attention is diffused to explore alternative image regions; otherwise, attention is reinforced to strengthen grounding in the same regions.
}
 \label{fig:architecture}
 \vspace{-5mm}
\end{figure*}

\section{Methodology}
%\zkn{Motivate before introducing the method}
We introduce a sink-triggered grounded decoding strategy that dynamically modulates multimodal self-attention based on grounding reliability during inference. Our approach is built on the observation that hallucinations frequently emerge around attention sink tokens - tokens that accumulate disproportionate attention while contributing little semantic content. SAGE intervenes precisely at these sink positions to assess whether the generation remains visually grounded, as depicted in Figure \ref{fig:architecture}.

At each detected sink token, we first extract key semantic concepts from the partially generated text. We then estimate visual grounding through two complementary signals: ($1$) the spatial self-attention map corresponding to the sink token projected onto visual tokens, and ($2$) gradient-based visual attribution (Grad-CAM) computed from the final layer of the vision encoder for the extracted concepts. To quantify grounding reliability, we compute a spatial overlap score between these two maps. This overlap serves as a real-time reliability indicator that guides attention modulation. High agreement indicates consistent grounding and prompts the model to reinforce attention within the same spatial regions, thereby strengthening reliable visual evidence. In contrast, low agreement signals potential misalignment and triggers the model to redistribute attention toward alternative image regions to mitigate emerging hallucinations.

% Based on this reliability score, SAGE adaptively modulates attention within the multimodal self-attention layers. When the overlap falls below a threshold, attention is broadened to explore alternative image regions and mitigate spurious fixation. Conversely, when the overlap is high, attention is sharpened to reinforce already grounded regions\zkn{These previous sentences seem redundant/can be comined with previous paragraph}.

The complete decoding procedure is summarized in Algorithm \ref{alg:sink_compact}. Through this selective, sink-triggered intervention, SAGE provides a training-free mechanism for improving visual grounding while preserving generative fluency. 
%\zkn{depending on what?}.
%\zkn{Need a more detailed overall description of the method that anticipates what you will be talking about later (sink detection, key concept extraction, grounding estimation, reliability estimation, and modulation. Put it all together to give the readers a global picture. }

\begin{algorithm}[t]
\caption{Sink-Aware Grounded Decoding with Attention Modulation
%\zk{where do you ref this? Might be good to have this in the beginning to give the holistic picture (or if not, an overview figure}
}
\label{alg:sink_compact}

\begin{algorithmic}[1]
\State Initialize sequence $S \leftarrow \emptyset$

\While{end-of-sequence not reached}
    \State Sample next token $t$ and append to $S$

    \If{$t$ is a designated sink token}
        \State Extract key concepts $C$ using POS tagging
        \State Obtain self-attention weights $A$ of token $t$ from middle decoder layers, restricted to image tokens
        \State Compute spatial attention map $i_1$ by projecting $A$ onto the image plane

        \State Compute visual grounding map 
        $i_2 \leftarrow \bigcup_{c \in C} \mathrm{GradCAM}(I, c)$
        
        \State Compute overlap score $o \leftarrow \mathrm{IoU}(i_1, i_2)$

        \If{$o > \tau$}
            \State Reinforce attention: $A \leftarrow \alpha_1 A$, where $\alpha_1 > 1$
        \Else
            \State Diffuse attention: $A \leftarrow \alpha_2 A$, where $\alpha_2 < 1$
        \EndIf
    \EndIf
\EndWhile

\State \Return generated sequence $S$

\end{algorithmic}
\vspace{-0.5mm}
\end{algorithm}

\noindent\textbf{Autoregressive Generation and Sink Detection.} To mitigate hallucination, we augment the standard autoregressive decoding procedure with a sink detection mechanism that operates continuously during decoding. Building upon our analysis in Sec.~\ref{sink_tok_analysis}, which demonstrates a strong correlation between attention sink tokens and hallucination events, we monitor generated tokens in real-time. Specifically, we classify a token as a potential attention sink if it belongs to a predefined set of punctuation marks and connective elements. This set includes boundary symbols (periods, commas, colons, semicolons, dashes, ellipses), coordinating conjunctions (e.g., "and", "or"), and similar structural tokens. While these tokens serve essential syntactic functions as phrase boundaries, they typically convey minimal semantic information, making them susceptible to accumulating disproportionate attention weights.%\zkn{Need to cite/explain your reasoning of connection to hallucinations/justify!}.
% Our framework employs standard autoregressive generation for decoding. Beginning with an empty output sequence, the model iteratively generates tokens conditioned on both the visual input and the previously generated sequence. Each newly generated token is appended to the output, and this process continues until termination criteria are satisfied.\zkn{This paragraph can be removed for space. I would just start with ``...we augmnet the standard autoregressive decoding procedure with a sink detection...'}

\label{sec:key_concept} \noindent \textbf{Key Concept Extraction.} Once an attention sink token is detected during decoding, we treat it as a semantic boundary: the tokens generated since the previous sink form a coherent descriptive segment. At each boundary, we extract key concepts from this newly generated segment i.e., tokens most likely to correspond to visual entities or attributes and use them both to assess the grounding of the current segment and to modulate subsequent decoding via attention control.

We operationalize this step using spaCy's part-of-speech (POS) \cite{ines_montani_2023_10009823} based concept identification strategy. Specifically, we apply POS tagging to the current output segment and extract tokens belonging to categories that typically carry high visual grounding potential. These include nouns, which usually denote objects or scene entities; adjectives, which describe visual attributes such as color, texture, or size; and compound noun phrases, which capture semantically complete units (e.g., “red sports car” or “wooden dining table”). By focusing on these categories, we ensure that grounding verification operates on concept-level semantic units rather than individual tokens.

This concept-level abstraction is crucial for two reasons. First, it reduces noise introduced by syntactic or functional tokens such as determiners, prepositions, or auxiliary verbs, which do not have direct visual correspondences. Second, it aligns the grounding process with how humans interpret visual descriptions - as collections of meaningful entities and attributes rather than isolated words. As a result, this step provides a robust set of candidate concepts that both underpin the grounding and attribution analysis of the current segment and guide the model’s future decoding behavior.

\noindent\textbf{Self-Attention Grounding Estimation.} After identifying key concepts from the output sequence before the sink token, the next step is to estimate the model’s internal visual grounding for these concepts. We achieve this by analyzing self-attention maps that capture the interaction between textual tokens and visual tokens within the multimodal transformer.

%\begin{itemize}

%\item  
% \noindent \textit{Attention Extraction.} In multimodal transformers, self-attention layers encode the alignment between generated text tokens and image tokens, providing a direct signal of grounding behavior. For each detected attention sink token, we extract its self-attention weights over image tokens to estimate the model’s spatial focus at that decoding step. We use self-attention maps from intermediate transformer layers and select the attention head that exhibits the strongest average attention mass over visual tokens for the sink position. This head most effectively captures semantic alignment between textual and visual representations. The exact layer index selection procedure is detailed in Sec.~\ref{sec:layer_analysis}.

% \zkn{Determined how? Usually details e.g. 14/15 might be in experiment section and here you can just say ``middle'' point to the details section}).
% \begin{figure}
%  \centering
%  \captionsetup{type=figure}
%  \includegraphics[width=0.48\linewidth]
%  {figures/fig4.1.pdf}
%  %\vspace{-20pt}
%  \caption{Layer-wise analysis showing the average IoU between spatial self-attention maps of ground-truth object tokens and their corresponding bounding boxes, alongside attention entropy. Results are averaged over $500$ images from MSCOCO.}
%  \label{fig:layer_analysis}
%  \vspace{-5mm}
% \end{figure}
\noindent \textit{Attention Extraction and Region Construction.}
In multimodal transformers, self-attention layers encode alignment between generated text tokens and image tokens, providing a direct signal of grounding. For each detected attention sink token, we extract its self-attention weights over image tokens from intermediate transformer layers, selecting the attention head with the highest average attention mass on visual tokens at the sink position (see Sec.~\ref{sec:layer_analysis} for layer selection). We then project these weights into the image plane to obtain a spatial grounding map, denoted $i_1$, which highlights the regions the model attends to when generating the corresponding textual content and serves as its internal estimate of the relevant visual evidence.
\begin{figure*}[!t]
\centering
\begin{subfigure}{0.65\linewidth}
    \includegraphics[width=\linewidth]{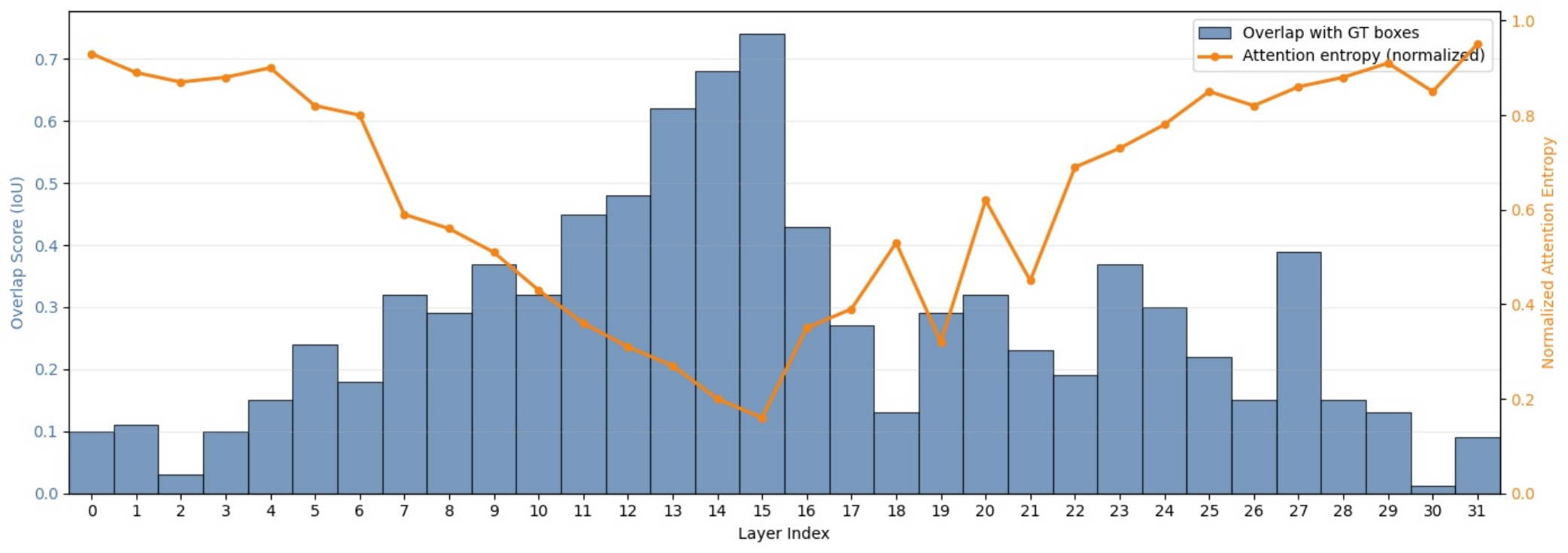}
 %\vspace{-20pt}

\end{subfigure}
\begin{subfigure}{0.33\linewidth}
    \includegraphics[width=\linewidth]{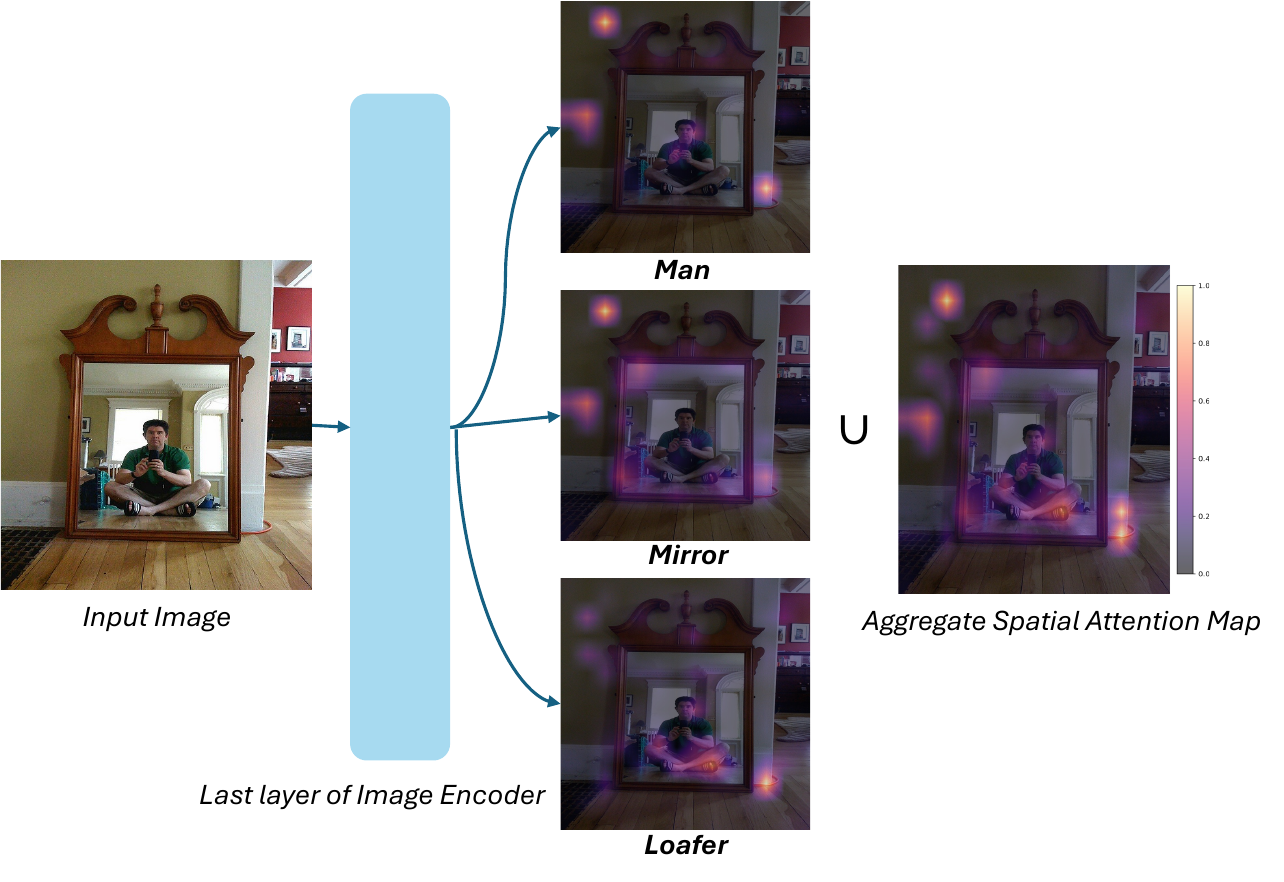}
\end{subfigure}
\caption{(a) Layer-wise analysis showing the average IoU between spatial self-attention maps of ground-truth object tokens and their corresponding bounding boxes, alongside attention entropy. Results are averaged over $500$ images from MSCOCO. (b) Grad-CAM visualizations from the final layer of the vision encoder for individual concepts and their union.}
\label{fig:analysis}
\vspace{-6mm}
\end{figure*}

\label{sec:gradcam}
\noindent\textbf{Gradient-Based Visual Attribution.}
To obtain an independent and causally grounded estimate of visual evidence, we complement self-attention analysis with gradient-based visual attribution. While self-attention reflects where the model attends during generation, it does not guarantee that those regions are causally responsible for predicting specific concepts. We therefore employ Grad-CAM~\cite{Selvaraju_2019} to derive an external grounding signal based on prediction sensitivity. We analyze the impact of the Grad-CAM component through an ablation study in \textbf{Appendix (Sec.~\ref{sec:supp_importance_gradcam})}.

For each extracted key concept $c \in \mathcal{C}$ identified prior to a sink trigger, we compute a concept-specific attribution map using Grad-CAM applied to the final-layer visual feature maps of the vision encoder \cite{zhang2024redundancyrelevanceinformationflow}. Importantly, the supervision signal does not originate from predefined visual classes or from the vision encoder itself. Instead, it is derived directly from the autoregressive decoder’s output logit corresponding to the concept token $c$.

\looseness=-1 Concretely, let $y_c$ denote the decoder logit associated with token $c$ in the generated sequence, and let $A^{\text{vision}}$ represent the final-layer visual feature activations. We backpropagate $\partial y_c / \partial A^{\text{vision}}$ through the multimodal transformer and compute Grad-CAM over $A^{\text{vision}}$, yielding a spatial attribution map that highlights the image regions most influential for predicting $c$ in the current context.

\looseness=-1 Since multiple concepts may be extracted before a sink token, we aggregate their attribution maps into a unified grounding representation as illustrated in Figure~\ref{fig:analysis}~b). Each concept-level Grad-CAM map is first thresholded to obtain a binary mask $i_2^{(c)}$, and we compute their pixel-wise union: $i_2 = \bigcup_{c \in \mathcal{C}} i_2^{(c)}$, where the union is implemented as a pixel-wise logical OR. The resulting region $i_2$ serves as an external grounding reference independent of attention dynamics. Unlike attention maps, which reflect internal information flow, gradient-based attribution provides a causal signal directly tied to prediction sensitivity, making it a more reliable reference for grounding verification.

\label{sec:reliability}\noindent\textbf{Grounding Reliability Estimation.} To assess the reliability of the model’s attention-based grounding signal, we evaluate the spatial consistency between the grounding induced by a detected sink token and the independently computed gradient-based attribution map for the extracted key concepts in $\mathcal{C}$. Specifically, for each identified sink token, %we derive its attention-based grounding map $i_1$ by projecting the sink token’s attention over visual tokens onto the image. W
we %then 
measure the spatial overlap between this sink-conditioned grounding map $i_1$ and the aggregated Grad-CAM-based map $i_2$ using the Intersection-over-Union (IoU) metric:  $o = \mathrm{IoU}(i_1, i_2).$
% \[
% o = \mathrm{IoU}(i_1, i_2).
% \]

This overlap score serves as a quantitative measure of spatial agreement between the sink token’s attention distribution and the gradient-based attribution of the extracted concepts. A high overlap value indicates that the regions currently emphasized by attention substantially coincide with the image areas that are causally influential for concept prediction. In this case, the model is already focusing on relevant visual evidence, and reinforcing attention within these regions can strengthen grounding consistency. Conversely, a low overlap indicates spatial misalignment between attention and attribution. This suggests that the current attention distribution does not adequately cover the regions most responsible for the predicted concepts, motivating a redistribution of attention toward alternative image areas.

Importantly, both scenarios arise at sink tokens, which represent structurally sensitive transition points in generation. The overlap score therefore does not classify tokens as hallucinated or grounded; rather, it determines the appropriate corrective action, i.e. whether to reinforce existing attention patterns or to diffuse attention to encourage broader visual exploration.

% By computing this agreement measure at sink tokens during decoding, our framework obtains a dynamic, context-aware signal that guides adaptive attention modulation. This allows SAGE to selectively reinforce reliable visual evidence or intervene when spatial misalignment emerges, thereby mitigating the propagation of hallucinated content.\zkn{Can be removed if needed}

\label{sec:modulation}\noindent\textbf{Sink-Aware Attention Modulation.}
Attention distributions in multimodal transformers are not inherently reliable indicators of visual grounding\cite{xiao2024efficientstreaminglanguagemodels}. At sink tokens, attention may either under-cover relevant image regions or concentrate on irrelevant areas, resulting in spatial misalignment with the visual evidence supporting the predicted concepts. To address this, we dynamically modulate the multimodal self-attention distribution based on the estimated grounding reliability score. Rather than treating attention weights as fixed, we adapt them according to their spatial agreement with an independent gradient-based reference: high overlap leads to reinforcement of the aligned regions, while low overlap triggers attention diffusion to encourage broader visual exploration.

% \vspace{-3mm}
\begin{figure*}[!t]
 \centering
 \captionsetup{type=figure}
 \includegraphics[width=0.95\textwidth]
 {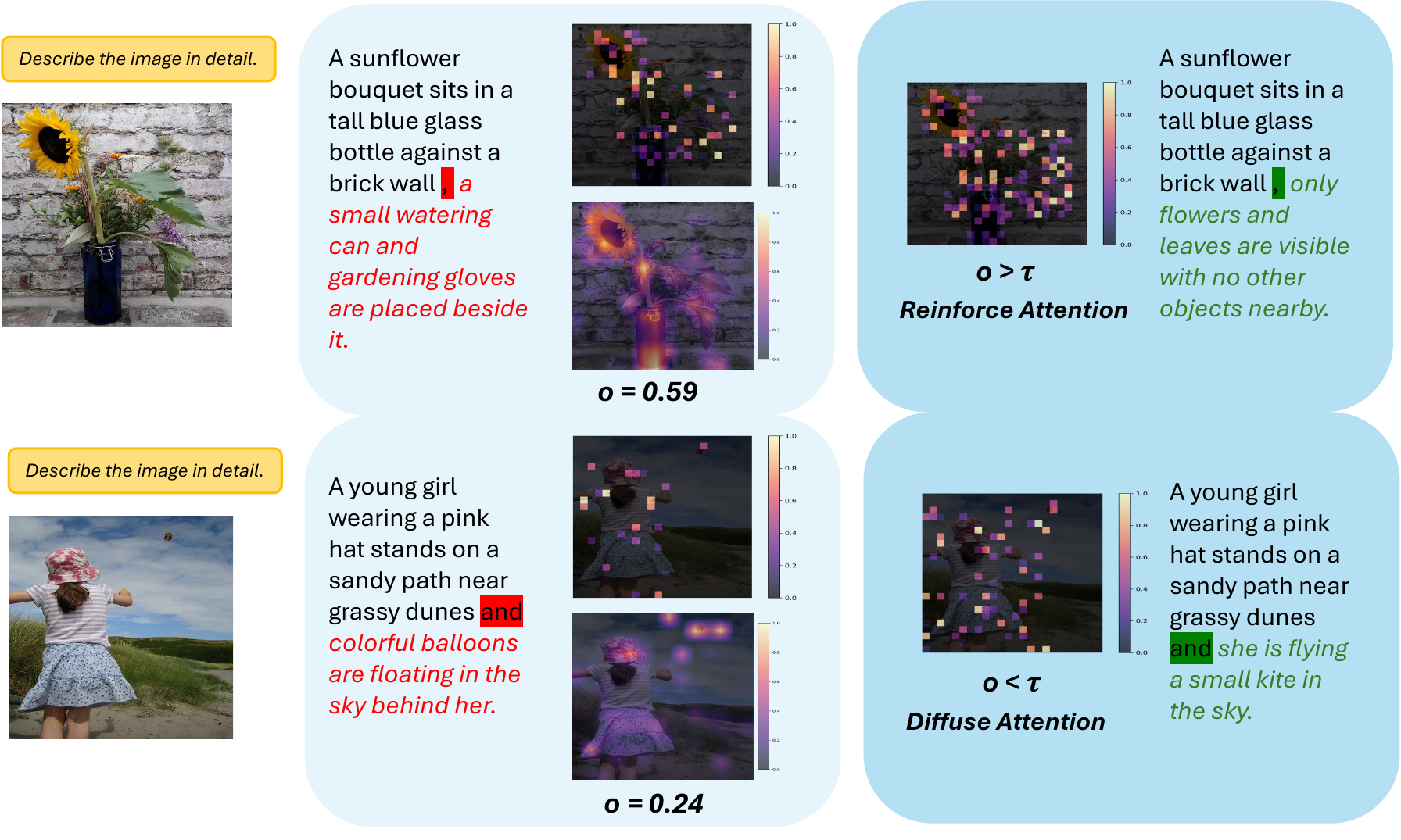}
 %\vspace{-20pt}
 \caption{Qualitative results of SAGE, including sink-token spatial attention, GradCAM-derived visual grounding maps, the computed overlap score $o$, and the modulated spatial attention maps. The sink token is highlighted in red.}
 \label{fig:qual_res}
 \vspace{-5mm}
\end{figure*}

Specifically, the overlap score between the attention-based and gradient-based grounding regions serves as a reliability signal. When this score exceeds a predefined threshold (empirically set to $0.5$ in our experiments), the attention distribution is considered spatially consistent with the gradient-based attribution, as depicted in Figure~\ref{fig:qual_res}. In this case, we reinforce the existing attention pattern by sharpening it, thereby increasing the relative weight assigned to the most highly attended visual tokens.
% \begin{equation}
% A \leftarrow \alpha_1 A, \quad \alpha_1 > 1.
% \end{equation}
This operation encourages the model to maintain strong focus on correctly grounded visual evidence and promotes faithful continuation of generation. Conversely, when the overlap falls below the threshold, the attention pattern is treated as unreliable. Such scenarios typically correspond to diffuse attention distributions or cases where the model focuses on visually irrelevant regions, both of which are strongly associated with hallucination. To mitigate this risk, we broaden the attention distribution by reducing its concentration.
% \begin{equation}
% A \leftarrow \alpha_2 A, \quad \alpha_2 < 1.
% \end{equation}
% \vspace{-2mm}
\[
A \leftarrow \alpha_1 A, \quad \alpha_1 > 1, \qquad
A \leftarrow \alpha_2 A, \quad \alpha_2 < 1.
\]
\vspace{-4mm}
% This softens over-confident attention peaks and allows the model to redistribute attention toward alternative visual regions that may contain relevant evidence. A detailed ablation analysis of the threshold selection and modulation strategy is provided in the supplementary material. 
% \zkn{Would be interesting to ablate, where you just sharpen (and ignore other case) and vice-versa. Also, I'm wondering if there is a simple method we can compare against in terms of \textit{when} you do this (sink tokens). For example, just doing it at periods or more frequently (which has downside of compute, but interesting to verify that sink token decision is important.}

This softens over-confident attention peaks and allows the model to redistribute attention toward alternative visual regions that may contain relevant evidence. A detailed ablation analysis of the threshold selection and modulation strategy along with schedules that trigger modulation at fixed periods instead of sink tokens is provided in the \textbf{Appendix (Sec. \ref{sec:supp_ablation_attention})}.

This adaptive modulation mechanism is applied selectively at attention sink points during decoding, enabling targeted intervention precisely when the model transitions between semantic units. As a result, the proposed approach provides an effective and lightweight strategy for steering attention toward reliable visual grounding without disrupting the overall auto-regressive generation process.

\section{Experiments}

\noindent \textbf{Models and Baselines.} We evaluate the effectiveness of our method across five representative multimodal large language models (MLLMs) that differ in their visual feature projection mechanisms. Specifically, we consider two models that employ simple linear projection layers, LLaVA-1.5-7b~\cite{liu2024improvedbaselinesvisualinstruction} and Shikra~\cite{chen2023shikraunleashingmultimodalllms}, which directly align visual and textual features and feed dense image tokens (typically 256 or 576 tokens) into the language model. We also include two Q-Former \cite{li2023blip2bootstrappinglanguageimagepretraining} based models, MiniGPT-4 \cite{zhu2023minigpt4enhancingvisionlanguageunderstanding} and InstructBLIP \cite{dai2023instructblipgeneralpurposevisionlanguagemodels}, where a Query Transformer compresses visual representations into a fixed set of learnable query tokens (e.g., 32 tokens), enabling more compact cross-modal alignment in the language embedding space.

All evaluated models utilize strong pretrained vision encoders, such as CLIP~\cite{radford2021learningtransferablevisualmodels} or EVA~\cite{fang2022evaexploringlimitsmasked}, together with pretrained large language models including LLaMA~\cite{touvron2023llamaopenefficientfoundation} or Vicuna~\cite{zheng2023judgingllmasajudgemtbenchchatbot}. In addition, we benchmark our method on Qwen2-VL-7B \cite{wang2024qwen2vlenhancingvisionlanguagemodels}, a recent state-of-the-art multimodal model featuring a dynamic-resolution vision transformer and tighter cross-modal integration through decoder cross-attention. Unlike projector or Q-Former-based designs, Qwen2-VL directly incorporates adaptively tokenized visual features into the language model, allowing efficient high-resolution processing while maintaining fine-grained visual grounding. This diverse set of architectures enables a comprehensive evaluation of our method across distinct multimodal integration paradigms.

We compare our method against several representative hallucination mitigation methods. OPERA~\cite{huang2024operaalleviatinghallucinationmultimodal} enhances beam search with an optimized decoding strategy. HALC~\cite{chen2024halcobjecthallucinationreduction} modifies global and local visual context to suppress hallucinations. EAH~\cite{zhang2024seeingclearlylayertwo} identifies vision sink tokens and adjusts shallow-layer attention. DML-LVLM~\cite{jiang2025devilsmiddlelayerslarge} redistributes visual attention across heads at inference time. GIFT~\cite{qi2025capturinggazeshiftsguidance} computes a holistic visual saliency map via gaze shifts and amplifies salient regions during decoding. LVLMs-Saliency~\cite{zhang-saliency} dynamically filters candidate tokens based on contextual saliency. We provide the details of our implementation setup in the \textbf{Appendix (Sec. \ref{sec:supp_implementation_details})}.

\noindent \textbf{Benchmarks.} We evaluate SAGE across a diverse set of hallucination detection benchmarks designed to measure both factual correctness and visual grounding in multimodal generation. These benchmarks assess complementary aspects of hallucination: whether generated descriptions mention objects not present in the image, and whether they sufficiently cover true visual content. In particular, we use the widely adopted CHAIR metric on MSCOCO to quantify object hallucination, and evaluate both CHAIR and Cover metrics on the AMBER benchmark to jointly assess hallucination and object coverage.

\noindent \textbf{CHAIR Evaluation on MSCOCO.} We first evaluate hallucination using the \textit{Caption Hallucination Assessment with Image Relevance (CHAIR)} \cite{rohrbach2019objecthallucinationimagecaptioning} metric on the MSCOCO dataset~\cite{lin2015microsoftcococommonobjects}. CHAIR is specifically designed to measure object hallucination in image captioning by comparing objects mentioned in generated descriptions with ground-truth object annotations. It computes the proportion of objects that appear in the caption but are absent from the image labels.

\begin{table}
\vspace{-3mm}
\centering
\caption{CHAIR$\downarrow$ results on MSCOCO for LVLMs with different decoding baselines aimed at hallucination mitigation. $C_S$ and $C_I$ denote $CHAIR_S$ and $CHAIR_I$, respectively. Blue text indicates the relative reduction (\%) achieved by SAGE over the best competing baseline (lower is better). All reported scores are averaged over generations with maximum lengths of 256 and 512 tokens.}
\label{tab:quant_mscoco}
\begin{tabularx}{\textwidth}{l*{10}{Y}}
\toprule
\multirow{2}{*}{Method} 
& \multicolumn{2}{c}{LLaVA-1.5} 
& \multicolumn{2}{c}{Shikra} 
& \multicolumn{2}{c}{MiniGPT-4} 
& \multicolumn{2}{c}{InstructBLIP} 
& \multicolumn{2}{c}{Qwen2-VL-7B} \\
\cline{2-11}
& $C_S\downarrow$ & $C_I\downarrow$ 
& $C_S\downarrow$ & $C_I\downarrow$ 
& $C_S\downarrow$ & $C_I\downarrow$ 
& $C_S\downarrow$ & $C_I\downarrow$ 
& $C_S\downarrow$ & $C_I\downarrow$ \\
\midrule
OPERA & 45.2 & 12.4 & 37.2 & 12.3 & 26.0 & 9.1 & 46.7 & 14.6 & 23.7 & 7.1 \\
HALC & 36.9 & 9.8 & 38.3 & 15.9 & 28.5 & 10.3 & 49.4 & 14.2 & 22.5 & 6.9 \\
EAH & 37.8 & 10.4 & 47.5 & 13.2 & 30.7 & 9.9 & 56.4 & \underline{9.9} & 22.4 & 6.5 \\
DML-LVLM & \underline{25.7} & \underline{7.2} & \underline{24.6} & 9.9 & \textbf{23.5} & 8.8 & \underline{32.5} & 13.8 & 25.7 & 7.2 \\
GIFT & 37.4 & 9.6 & 33.5 & 11.7 & 29.3 & 10.2 & 38.8 & 11.9 & 22.5 & 8.1 \\
LVLMs-Saliency & 36.2 & 8.8 & 31.6 & 10.2 & 27.4 & 9.1 & 35.3 & 10.2 & \textbf{20.1} & \textbf{5.4} \\
\midrule
SAGE (Ours) 
& \textbf{21.3} {\color{blue}\tiny(+17.1\%)} 
& \textbf{5.2} {\color{blue}\tiny(+27.8\%)} 
& \textbf{24.3} {\color{blue}\tiny(+1.2\%)} 
& \textbf{8.7} {\color{blue}\tiny(+12.1\%)} 
& \underline{23.8} {\color{blue}\tiny(+1.3\%)} 
& \textbf{8.2} {\color{blue}\tiny(+6.8\%)} 
& \textbf{25.7} {\color{blue}\tiny(+20.9\%)} 
& \textbf{9.4} {\color{blue}\tiny(+5.1\%)} 
& \underline{21.1} {\color{blue}\tiny(+4.9\%)} 
& \underline{5.9} {\color{blue}\tiny(+9.3\%)} \\
\bottomrule
\end{tabularx}%
% \vspace{-3mm}
\end{table}

\noindent CHAIR provides two complementary variants: per-sentence, or what fraction of sentences include a hallucinated object (denoted as $CHAIR_S$) and per-instance, or what fraction of object instances are hallucinated (denoted as $CHAIR_I$).
% \begin{itemize}
    
\vspace{-4mm}
\[
C_I = \frac{|\{\text{hallucinated objects}\}|}{|\{\text{all mentioned objects}\}|} \qquad
C_S = \frac{|\{\text{captions with hallucinated objects}\}|}{|\{\text{all captions}\}|}
\]

% \begin{equation}
% C_I = \frac{|\{\text{captions with hallucinated objects}\}|}{|\{\text{all captions}\}|}.
% \end{equation}

% \end{itemize}
% We conduct CHAIR evaluation on the MSCOCO dataset, which contains over 300{,}000 images annotated with 80 object categories. Following standard practice, we randomly sample 500 images from the COCO-2014 validation split and generate descriptions using the prompt \textit{``Please describe this image in detail.''} To ensure fair comparison, we fix the maximum generation length to 256 and 512 tokens across all baselines and compute the average of the two setups. As evident from Table \ref{tab:quant_mscoco}, results demonstrate that SAGE consistently achieves lower $CHAIR_S$ and $CHAIR_I$ scores than existing decoding methods, indicating significantly reduced hallucination. \zkn{Some of these details (prompt etc.) can be moved to supplementary, and need a bit more on description of results}

We conduct CHAIR evaluation on the MSCOCO dataset, which contains over 300{,}000 images annotated with 80 object categories; full details of the sampling protocol, prompts, and decoding setup are provided in the \textbf{Appendix (Sec. \ref{sec:supp_chair_mscoco})}. As evident from Table~\ref{tab:quant_mscoco}, SAGE consistently achieves lower $CHAIR_S$ and $CHAIR_I$ scores than all competing decoding methods across the five LVLM backbones, indicating substantially reduced hallucination. For example, on LLaVA-1.5-7b SAGE reduces $CHAIR_S$ from 25.7 to 21.3 and $CHAIR_I$ from 7.2 to 5.2 (17.1\% and 27.8\% relative reductions) with consistent gains on other architectures. The average relative improvement on MSCOCO is 10.65\% across vlm backbones.

\noindent \textbf{CHAIR and Cover Evaluation on AMBER.} To further evaluate captioning quality beyond hallucination detection, we conduct experiments on the AMBER \cite{wang2024amberllmfreemultidimensionalbenchmark} benchmark, a multi-dimensional evaluation suite designed to assess multimodal generation without reliance on external language models. AMBER enables simultaneous measurement of hallucination and object coverage, providing a more comprehensive assessment of grounding performance.

\begin{table}
\vspace{-3mm}
  \centering
  \caption{CHAIR$\downarrow$ and Cover$\uparrow$ results on AMBER for LVLMs with different decoding baselines aimed at hallucination mitigation. $C_I$ and $C_R$ denote $CHAIR_I$ and Cover, respectively. Blue text indicates the relative improvement (\%) achieved by SAGE over the best competing baseline (lower is better for $C_I$, higher is better for $C_R$). All reported scores are averaged over generations with maximum lengths of 256 and 512 tokens.}
  \label{tab:quant_amber}
  \resizebox{\textwidth}{!}{%
  \begin{tabularx}{\textwidth}{l*{10}{Y}}
    \toprule
     \multirow{2}{*}{Method} 
     & \multicolumn{2}{c}{LLaVA-1.5} 
     & \multicolumn{2}{c}{Shikra} 
     & \multicolumn{2}{c}{MiniGPT-4} 
     & \multicolumn{2}{c}{InstructBLIP} 
     & \multicolumn{2}{c}{Qwen2-VL-7B} \\
     \cline{2-11}
     & $C_I\downarrow$ & $C_R\uparrow$ 
     & $C_I\downarrow$ & $C_R\uparrow$ 
     & $C_I\downarrow$ & $C_R\uparrow$ 
     & $C_I\downarrow$ & $C_R\uparrow$ 
     & $C_I\downarrow$ & $C_R\uparrow$ \\
    \midrule
    OPERA  
    & 12.1 & 45.3 & 11.2 & 42.6 & 10.3 & 50.4 & 12.2 & 48.7 & 8.8 & 52.4 \\

    HALC 
    & 9.7 & 51.7 & 16.2 & 39.6 & 11.1 & 53.6 & 13.1 & 49.3 & 7.8 & 54.7 \\

    EAH
    & 11.3 & 48.4 & 12.6 & 52.3 & 10.6 & 54.5 & 10.7 & 49.1 & 7.1 & 55.2 \\

    DML-LVLM
    & \underline{7.8} & 52.6 & \underline{9.2} & 46.7 & 8.9 & 47.4 & 13.6 & 44.3 & 8.1 & 51.6 \\

    GIFT
    & 8.9 & 50.6 & 10.2 & 47.2 & 9.6 & 41.8 & 11.2 & 39.4 & \underline{6.3} & 58.7 \\

    LVLMs-Saliency
    & 7.9 & \underline{51.3} 
    & 9.6 & \underline{50.8} 
    & \underline{8.4} & \textbf{56.7} 
    & \underline{9.6} & 46.9 
    & \underline{6.7} & \textbf{64.9} \\

    \midrule
    SAGE (Ours) 
    & \textbf{6.9} {\color{blue}\tiny(+11.5\%)} 
    & \textbf{57.2} {\color{blue}\tiny(+0.9\%)} 

    & \textbf{8.3} {\color{blue}\tiny(+9.8\%)} 
    & \textbf{53.4} {\color{blue}\tiny(+2.1\%)} 

    & \textbf{7.6} {\color{blue}\tiny(+9.5\%)} 
    & \underline{53.2} {\color{blue}\tiny(+2.9\%)} 

    & \textbf{8.2} {\color{blue}\tiny(+14.6\%)} 
    & \textbf{55.7} {\color{blue}\tiny(+13.1\%)} 

    & \textbf{6.1} {\color{blue}\tiny(+3.2\%)} 
    & \underline{61.5} {\color{blue}\tiny(+4.3\%)} \\
    \bottomrule
  \end{tabularx}%
  }
  % \vspace{-6mm}
\end{table}

\looseness=-1 In addition to CHAIR, AMBER introduces the \textbf{Cover} metric, which measures how well generated descriptions capture true visual content. Specifically, Cover quantifies the proportion of ground-truth objects that are correctly mentioned in the response: $\text{Cover}(R) = \frac{|R'_{\text{obj}} \cap A_{\text{obj}}|}{|A_{\text{obj}}|}$, where $R'_{\text{obj}}$ denotes objects extracted from the generated response, and $A_{\text{obj}}$ denotes ground-truth annotated objects. An ideal model achieves low hallucination while maintaining high coverage of image content.

% For evaluation, we use the standard generative prompt \textit{``Describe this image in detail.''} and randomly sample 500 instances from the 1{,}004 available datapoints in AMBER. We report both $CHAIR_I$ and Cover scores to assess the trade-off between hallucination reduction and descriptive completeness. As observed in Table \ref{tab:quant_amber}, results show that SAGE not only reduces hallucination but also preserves strong object coverage, demonstrating its ability to improve factual grounding without sacrificing descriptive quality.\zkn{Some of these details (prompt etc.) can be moved to supplementary, and need a bit more on description of results}

\looseness=-1 For evaluation, we use the standard generative prompt and randomly sample 500 instances from the 1,004 available datapoints in AMBER; full details are in the \textbf{Appendix (Sec. \ref{sec:supp_chair_amber})}. We report both $CHAIR_I$ and Cover scores to assess the trade-off between hallucination reduction and descriptive completeness. As observed in Table \ref{tab:quant_amber}, SAGE not only reduces hallucination but also preserves or improves object coverage, demonstrating its ability to improve factual grounding without sacrificing descriptive quality. For instance, on LLaVA-1.5, SAGE achieves the best $C_I$ score of 6.9 (a 11.5\% relative reduction over the next-best baseline) while also attaining the highest Cover score of 57.2, showcasing a strong balance between faithfulness and completeness across all five LVLM backbones.
  
\textbf{Gemini-3-Pro Based Human-Aligned Evaluation.}
To assess hallucinations from a human-aligned perspective, we employ a strong multimodal judge, Gemini~3~Pro~\cite{gemini3_2025}, following~\cite{Yin_2024}. We randomly sample 100 images from the MSCOCO validation set and generate detailed captions using the prompt \emph{``Please describe this image in detail.''} for GIFT~\cite{qi2025capturinggazeshiftsguidance}, LVLMs-Saliency~\cite{zhang-saliency}, and SAGE on LLaVA-1.5 and Qwen2-VL-7B. For evaluation, Gemini~3~Pro is provided with the input image and three candidate responses simultaneously for each model, and is asked to assign comparative scores (0-10) along two dimensions: \emph{accuracy} and \emph{detailedness}. The prompt explicitly instructs the judge to penalize hallucinations, including non-existent objects and incorrect attributes (e.g., color, position, or relationships), and is designed to be order-invariant to mitigate positional bias. This protocol serves as a proxy for human evaluation given Gemini~3~Pro’s strong multimodal reasoning capabilities. The evaluation prompt is provided in the \textbf{Appendix (Sec. \ref{sec:supp_gemini_eval})}. As shown in Table~\ref{tab:quant_gemini}, SAGE consistently achieves higher scores, indicating improved factual correctness while maintaining rich descriptions.

We provide a detailed runtime analysis and additional qualitative results in the \textbf{Appendix (Secs.\ref{sec:supp_runtime} and \ref{sec:supp_qual})}.

\begin{table}
% \vspace{-3mm}
  \centering
  \caption{Gemini~3~Pro-based evaluation results. $A$ stands for 
  \emph{accuracy} and $D$ stands for \emph{detailedness}.} 
  \label{tab:quant_gemini}
  % \vspace{-3mm}
  
  \begin{tabularx}{0.6\textwidth}{l*{4}{Y}}
    \toprule
     \multirow{2}{*}{Method} & \multicolumn{2}{c}{LLaVA-1.5} &  \multicolumn{2}{c}{Qwen2-vl-7b} \\
     \cline{2-5}
     \multirow{2}{*}{} & \multicolumn{1}{c}{\textit{$A$}} & \multicolumn{1}{c}{\textit{$D$}} & \multicolumn{1}{c}{\textit{$A$}} & \multicolumn{1}{c}{\textit{$D$}} \\
    \midrule
    GIFT \cite{qi2025capturinggazeshiftsguidance} & $4.3$ & $4.7$& $6.5$ & $6.8$ \\
    LVLMs-Saliency \cite{zhang-saliency} & $5.6$ & $5.9$ & $6.1$ & $5.8$ \\
    SAGE(Ours) & $\textbf{6.9}$ & $\textbf{6.6}$ & $\textbf{8.2}$ & $\textbf{8.1}$ \\
    \bottomrule
  \end{tabularx}%

   \vspace{-3mm}
\end{table}

\noindent\textbf{Transformer Layer Analysis.} \label{sec:layer_analysis}
% \zkn{Could put in experiments (and point to that section here, but don't have to. We can see what makes sense in the end. It might confuse reviewers to think that you're using ground truth in your method. Alternatively, make a strong subheading saying this is analysis only NOT the method.}
To justify our choice of layers for extracting attention maps, we conduct an analysis-only study using two complementary metrics: \textit{object localization accuracy} and \textit{attention entropy}. We first extract the ground-truth concepts from the generated output sequence and use the corresponding self-attention weights. For localization, these concept-specific self-attention weights at each layer are projected into spatial heatmaps over the image to obtain attention-based grounding regions. These regions are then compared against ground-truth bounding boxes using Intersection-over-Union (IoU).  In parallel, we quantify attention concentration by computing the entropy of the normalized attention distribution.
% \vspace{-2mm}
\[
\mathrm{IoU}(l) = \frac{\lvert R_{\text{attn}}^{(l)} \cap R_{\text{GT}} \rvert}{\lvert R_{\text{attn}}^{(l)} \cup R_{\text{GT}} \rvert}, \qquad
H(l) = - \sum_i A_i^{(l)} \log A_i^{(l)}
\]
where $R_{\text{attn}}^{(l)}$ denotes the attention-derived region at layer $l$, and $R_{\text{GT}}$ denotes the ground-truth object regions and $A_i^{(l)}$ is the attention weight assigned to image token $i$ at layer $l$. Lower entropy corresponds to more spatially focused attention. Results are averaged over $500$ images randomly sampled from the MSCOCO validation set.

 As shown in Figure~\ref{fig:analysis}~a), both analysis reveal a consistent trend: early layers yield low IoU and high entropy, indicative of coarse, globally distributed processing; middle layers attain the highest IoU and lowest entropy, reflecting the most precise and semantically aligned grounding; and later layers show decreasing IoU and rising entropy, suggesting a shift toward predominantly linguistic processing. These observations support our use of middle transformer layers as providing the most informative signals for visual-semantic alignment.
 % \zkn{Potential move to supplementary and just reference here instead. In any case, I wouldn't start with this for experiments but rather move to after main results. }
 
\vspace{-3mm}
\section{Conclusion}

% We introduced \textbf{SAGE}, a sink-aware grounded decoding strategy for mitigating hallucinations in large vision–language models. SAGE treats attention sink tokens as semantic checkpoints during autoregressive generation, using them to trigger concept-level grounding checks and dynamic modulation of attention over visual tokens. By integrating internal self-attention signals with gradient-based visual attribution, our method offers a training-free, interpretable mechanism for strengthening visual grounding at inference time. Across multiple benchmarks, SAGE consistently reduces hallucinations while preserving the descriptive quality of generated outputs. 
% A key limitation of SAGE is its reliance on the occurrence of sink tokens as intervention points. Consequently, it is less applicable to tasks with very short outputs (e.g., single-word or short-phrase answers and other highly discriminative settings), where such structural tokens may be rare or absent. Extending grounded decoding to these regimes is an important direction for future work. We also provide more failure cases in \textbf{Supplementary material}. 

We introduced \textbf{SAGE}, a sink-aware grounded decoding strategy for mitigating hallucinations in large vision–language models. SAGE treats attention sink tokens as semantic checkpoints during autoregressive generation, triggering concept-level grounding checks and dynamic modulation of attention over visual tokens. By combining internal self-attention signals with gradient-based visual attribution, our method provides a training-free and interpretable mechanism for improving visual grounding at inference time. Experiments across multiple benchmarks show that SAGE consistently reduces hallucinations while preserving descriptive quality.

A limitation of SAGE is its reliance on sink tokens as intervention points. As a result, it is less effective for tasks with very short outputs (e.g., single-word or short-phrase answers), where such tokens may rarely occur. Extending grounded decoding to these settings is an important direction for future work. Additional failure cases are provided in the \textbf{Appendix (Sec. \ref{sec:supp_failure})}.

% We presented \textbf{SAGE}, a sink-aware grounded decoding strategy for mitigating hallucinations in large vision–language models. SAGE treats attention sink tokens as semantic checkpoints during autoregressive generation, triggering concept-level grounding verification and adaptive attention modulation over visual tokens. By combining internal self-attention signals with gradient-based visual attribution, SAGE provides a training-free and interpretable mechanism to strengthen visual grounding at inference time. Experiments across multiple benchmarks show that SAGE consistently reduces hallucinations while preserving descriptive quality. A limitation of our method is its reliance on sink tokens as intervention points, making it less effective for tasks with very short outputs (e.g., single-word answers). Extending grounded decoding to such settings remains an important direction for future work.

\section{Appendix}

\subsection{Implementation Details}
\label{sec:supp_implementation_details}
We evaluate our method on two widely used benchmarks for image captioning hallucination: CHAIR~\cite{rohrbach2019objecthallucinationimagecaptioning} on the MSCOCO dataset~\cite{lin2015microsoftcococommonobjects}, and AMBER~\cite{wang2024amberllmfreemultidimensionalbenchmark}. Key semantic concepts are extracted from the generated sequence using spaCy-based part-of-speech (POS) tagging~\cite{ines_montani_2023_10009823}. We define sink tokens as punctuation marks and lightweight function tokens that frequently accumulate attention during decoding. The set used in our experiments includes punctuation tokens (., ,, :, ;, !, ?, -, --, ...) and conjunctions (and, or, but, so, yet).

The hyperparameters $\alpha_1$, $\alpha_2$, and $\tau$ are tuned on small validation subsets consisting of 200 samples from each dataset. We select $\alpha_1$ from [${0.3, 0.5, 0.6, 0.8}$]. For LLaVA-1.5~\cite{liu2023visualinstructiontuning}, Shikra~\cite{chen2023shikraunleashingmultimodalllms}, MiniGPT-4~\cite{zhu2023minigpt4enhancingvisionlanguageunderstanding}, and InstructBLIP~\cite{dai2023instructblipgeneralpurposevisionlanguagemodels}, $\alpha_1$ is set to $0.6$, while for Qwen2-VL-7B~\cite{wang2024qwen2vlenhancingvisionlanguagemodels} it is set to $0.3$. Similarly, $\alpha_2$ is chosen from [${1.2, 1.5, 1.8, 2.0}$] and set to $1.2$ for Qwen2-VL-7B and $1.8$ for the remaining models. The overlap threshold $\tau$ is set to $0.5$. Detailed ablation results are provided in Table~\ref{tab:ablation_hyperparams}. Overall, we observe that SAGE remains robust to moderate variations of these hyperparameters across benchmarks.

Following~\cite{zhang2024redundancyrelevanceinformationflow}, Grad-CAM is computed using the feature maps from the final layer of the vision encoder, as they capture rich high-level semantic information while retaining spatial structure.

\subsection{Importance of Grad-CAM}
\label{sec:supp_importance_gradcam}
To highlight the role of Grad-CAM~\cite{Selvaraju_2019}, we conduct an ablation study where the gradient-based attribution component is removed. Instead of computing the overlap between Grad-CAM maps from the final layer of the image encoder and the self-attention maps from intermediate LLM decoder layers, we rely solely on the spatial attention maps and apply a fixed overlap threshold of 0.5 with respect to the image area to decide whether to reinforce or diffuse attention. As shown in Table~\ref{tab:supple_ablation_mscoco}, the full method with Grad-CAM consistently outperforms this variant. This demonstrates that Grad-CAM provides valuable concept-level visual attribution from the vision encoder, enabling more reliable grounding signals than attention maps alone.

\begin{table}
% \vspace{-3mm}
\centering
\caption{CHAIR$\downarrow$ results on MSCOCO for SAGE with and without Grad-CAM. $C_S$ and $C_I$ denote $CHAIR_S$ and $CHAIR_I$, respectively. All reported scores are averaged over generations with maximum lengths of 256 and 512 tokens.}
\label{tab:supple_ablation_mscoco}

\small
\setlength{\tabcolsep}{2pt}
\resizebox{0.9\textwidth}{!}{%
\begin{tabular}{lcccccccccc}
\toprule
\multirow{2}{*}{Method} 
& \multicolumn{2}{c}{LLaVA-1.5} 
& \multicolumn{2}{c}{Shikra} 
& \multicolumn{2}{c}{MiniGPT-4} 
& \multicolumn{2}{c}{InstructBLIP} 
& \multicolumn{2}{c}{Qwen2-VL-7B} \\
\cline{2-11}
& $C_S\downarrow$ & $C_I\downarrow$
& $C_S\downarrow$ & $C_I\downarrow$
& $C_S\downarrow$ & $C_I\downarrow$
& $C_S\downarrow$ & $C_I\downarrow$
& $C_S\downarrow$ & $C_I\downarrow$ \\
\midrule
SAGE w/o Grad-CAM 
& 47.4 & 13.2 & 50.8 & 13.9 & 49.6 & 14.6 & 54.2 & 13.7 & 38.3 & 11.2 \\

SAGE (Ours) 
& \textbf{21.3} & \textbf{5.2}
& \textbf{24.3} & \textbf{8.7}
& \textbf{23.8} & \textbf{8.2}
& \textbf{25.7} & \textbf{9.4}
& \textbf{21.1} & \textbf{5.9} \\
\bottomrule
\end{tabular}
}
\end{table}

\subsection{Ablation Analysis for Attention Modulation}
\label{sec:supp_ablation_attention}

\begin{table}
% \vspace{-3mm}
\centering
\caption{Ablation study of SAGE hyperparameters $\alpha_1$, $\alpha_2$, and $\tau$ on MSCOCO. We report $C_S$, $C_I$, and $F1$ for LLaVA-1.5b and Qwen2-vl-7b. Lower values are better for $C_S$ and $C_I$, while higher is better for $F1$. All reported scores are generated with maximum lengths of 512 tokens.}
\label{tab:ablation_hyperparams}
\begin{tabularx}{\textwidth}{ccc*{6}{Y}}
\toprule
\multicolumn{3}{c}{Hyperparameters} 
& \multicolumn{3}{c}{LLaVA-1.5} 
& \multicolumn{3}{c}{Qwen2-VL-7B} \\
\cmidrule(lr){1-3} \cmidrule(lr){4-6} \cmidrule(lr){7-9}
$\alpha_1$ & $\alpha_2$ & $\tau$ 
& $C_S\downarrow$ & $C_I\downarrow$ & $F1\uparrow$
& $C_S\downarrow$ & $C_I\downarrow$ & $F1\uparrow$ \\
\midrule

\multicolumn{9}{c}{\textit{Reinforce only ($\alpha_1=0$)}} \\
0.0 & 1.2 & 0.5 & 56.7 & 14.7 & 50.6 & 49.9 & 13.8 & 58.9 \\
0.0 & 1.5 & 0.5 & 57.2 & 14.8 & 47.7 & 50.3 & 13.9 & 56.7 \\
0.0 & 1.8 & 0.5 & 57.9 & 15.3 & 48.0 & 53.7 & 14.2 & 55.2 \\
0.0 & 2.0 & 0.5 & 59.3 & 15.2 & 48.7 & 54.2 & 14.3 & 53.2 \\

\midrule
\multicolumn{9}{c}{\textit{Diffuse only ($\alpha_2=0$)}} \\
0.3 & 0.0 & 0.5 & 55.2 & 14.8 & 48.2 & 50.3 & 13.6 & 54.2 \\
0.6 & 0.0 & 0.5 & 57.4 & 14.9 & 46.4 & 48.5 & 13.4 & 52.7 \\
0.8 & 0.0 & 0.5 & 56.1 & 15.3 & 47.8 & 48.9 & 12.9 & 53.6 \\
1.0 & 0.0 & 0.5 & 55.8 & 14.9 & 49.3 & 46.7 & 13.2 & 55.4 \\

\midrule
\multicolumn{9}{c}{\textit{Both modulations active}} \\

0.3 & 1.8 & 0.5 & 39.7 & 11.1 & 63.7 & 41.7 & 10.1 & 69.2 \\
0.3 & 1.2 & 0.5 & 32.1 & 8.8 & 76.2 & \textbf{27.3} & \textbf{6.4} & \textbf{87.4} \\
0.5 & 1.5 & 0.5 & 35.2 & 12.5 & 67.2 & 39.1 & 9.7 & 71.6 \\
0.6 & 1.2 & 0.5 & 37.9 & 10.7 & 72.4 & 41.4 & 9.1 & 75.3 \\
0.6 & 1.8 & 0.5 & \textbf{29.7} & \textbf{7.5} & \textbf{84.0} & 36.2 & 9.4 & 79.3 \\
0.8 & 1.0 & 0.5 & 38.2 & 11.3 & 83.1 & 37.2 & 10.4 & 70.3 \\

\bottomrule
\end{tabularx}
% \vspace{-3mm}
\end{table}

We perform an ablation analysis on the hyperparameters $\alpha_1$, $\alpha_2$, and $\tau$, and additionally study the role of sink tokens by applying attention modulation at fixed intervals (every 10 decoding steps) instead of using sink-triggered checkpoints. Table~\ref{tab:ablation_hyperparams} reports the results on MSCOCO for LLaVA-1.5 and Qwen2-VL-7B.

As observed in Table~\ref{tab:ablation_hyperparams}, when only the broadening mechanism is active ($\alpha_2=0$), attention is diffused across image regions, which weakens grounding signals and leads to higher hallucination rates and lower F1 scores. Conversely, enabling only the sharpening mechanism ($\alpha_1=0$) concentrates attention on previously attended regions but lacks the ability to recover from incorrect focus, resulting in limited improvements. The best performance is achieved when both mechanisms are active, allowing the model to adaptively reinforce reliable regions while redistributing attention when grounding is uncertain.

Furthermore, as shown in Table~\ref{tab:supple_ablation_modulation}, replacing sink-triggered modulation with fixed-step intervention (every 10 decoding steps) leads to noticeably worse performance across all metrics. This degradation indicates that arbitrary intervention points are less effective, as they may occur when the model is already grounded or before grounding failures emerge. In contrast, sink tokens naturally correspond to transition points in generation where the model shifts between semantic concepts, making them reliable checkpoints for grounding verification and targeted attention modulation.

\begin{table}
% \vspace{-3mm}
\centering
\caption{Ablation study of the modulation strategy on MSCOCO. We compare periodic modulation applied every 10 decoding steps with our sink-triggered modulation strategy. $C_S$ and $C_I$ denote $CHAIR_S$ and $CHAIR_I$, respectively. All reported scores are averaged over generations with maximum lengths of 256 and 512 tokens.}
\label{tab:supple_ablation_modulation}

\small
\setlength{\tabcolsep}{2pt}
\resizebox{\textwidth}{!}{%
\begin{tabular}{lcccccccccc}
\toprule
\multirow{2}{*}{Method} 
& \multicolumn{2}{c}{LLaVA-1.5} 
& \multicolumn{2}{c}{Shikra} 
& \multicolumn{2}{c}{MiniGPT-4} 
& \multicolumn{2}{c}{InstructBLIP} 
& \multicolumn{2}{c}{Qwen2-VL-7B} \\
\cline{2-11}
& $C_S\downarrow$ & $C_I\downarrow$
& $C_S\downarrow$ & $C_I\downarrow$
& $C_S\downarrow$ & $C_I\downarrow$
& $C_S\downarrow$ & $C_I\downarrow$
& $C_S\downarrow$ & $C_I\downarrow$ \\
\midrule

Periodic Modulation (every 10 tokens) 
& 39.4 & 12.5 
& 41.3 & 12.9 
& 39.9 & 13.1 
& 38.2 & 11.7 
& 34.6 & 10.8 \\

Sink-triggered Modulation (SAGE) 
& \textbf{21.3} & \textbf{5.2}
& \textbf{24.3} & \textbf{8.7}
& \textbf{23.8} & \textbf{8.2}
& \textbf{25.7} & \textbf{9.4}
& \textbf{21.1} & \textbf{5.9} \\

\bottomrule
\end{tabular}
}
\end{table}

\subsection{CHAIR Evaluation on MSCOCO}
\label{sec:supp_chair_mscoco}
We evaluate hallucination using the CHAIR metric~\cite{rohrbach2019objecthallucinationimagecaptioning} on the MSCOCO dataset~\cite{lin2015microsoftcococommonobjects}, which contains over 300{,}000 images annotated with 80 object categories. Following standard practice, we randomly sample 500 images from the COCO-2014 validation split. For each image, captions are generated using the prompt \textit{``Please describe this image in detail.''} under an autoregressive decoding setup. To ensure fair comparison across methods, we keep all decoding settings identical and evaluate generations with maximum token lengths of 256 and 512. The reported results correspond to the average performance across these two generation settings. Hallucination is quantified using both $CHAIR_S$ (sentence-level hallucination rate) and $CHAIR_I$ (instance-level hallucination rate), which measure the presence of objects in the generated caption that are not present in the ground-truth annotations.

\subsection{CHAIR and COVER Evaluation on AMBER}
\label{sec:supp_chair_amber}
For evaluation on AMBER~\cite{wang2024amberllmfreemultidimensionalbenchmark}, we follow the standard generative protocol and use the prompt \textit{``Describe this image in detail.''}. From the 1{,}004 available datapoints in the benchmark, we randomly sample 500 images for evaluation. To assess both hallucination and descriptive completeness, we report the $CHAIR_I$ metric alongside the Cover score. While $CHAIR_I$ measures the rate of hallucinated object instances that are not present in the image, Cover evaluates the extent to which objects present in the image are correctly mentioned in the generated caption. This combination allows us to analyze the trade-off between reducing hallucinations and maintaining comprehensive image descriptions.

\subsection{Gemini-3-Pro Evaluation}
\label{sec:supp_gemini_eval}

Below is the system prompt given to Gemini-3-Pro~\cite{gemini3_2025} for evaluation: 

% \begin{center}
% \textbf{Prompt 1: Gemini-3-Pro Evaluation Prompt}
% \end{center}

\noindent\rule{\linewidth}{0.1pt}

\begin{algorithmic}

\State You are an expert evaluator for multimodal AI systems. Your task is to assess the quality of image descriptions generated by multiple AI assistants. The goal is to determine which descriptions best reflect the actual visual content of the image. You must carefully compare the image with each response and evaluate them according to two criteria: \textit{Accuracy} and \textit{Detailedness}.

\State \textbf{Hallucination Definition:}
\State A hallucination refers to any description that is inconsistent with the image. This includes:
\State \quad • Mentioning objects not present in the image
\State \quad • Incorrect object counts
\State \quad • Incorrect attributes (color, size, shape)
\State \quad • Incorrect spatial relationships
\State \quad • Unsupported actions or interactions

\State \textbf{Evaluation Criteria}

\State \quad 1. \textbf{Accuracy:} Evaluate how factually correct the description is with respect to the image. Descriptions that contain hallucinations or incorrect attributes should receive lower scores.

\State \quad 2. \textbf{Detailedness:} Evaluate whether the description provides meaningful and useful visual details about the scene. Valid details include correct objects, attributes, actions, and spatial relationships that are clearly supported by the image. Hallucinated information must NOT be counted as valid detail.

\State \textbf{Scoring Rules}

\State \quad • Assign a score from 0 to 10 for each assistant for both criteria.
\State \quad • Higher scores indicate better performance.
\State \quad • Accuracy should take priority over detailedness.
\State \quad • Do not reward verbosity unless the content is correct and supported by the image.
\State \quad • Evaluate each assistant independently based on the same standards.

\State \textbf{Input Format}

\State You will be given an image and three candidate responses.

\State \quad [Assistant 1] \{Response of Assistant 1\} [End of Assistant 1]
\State \quad [Assistant 2] \{Response of Assistant 2\} [End of Assistant 2]
\State \quad [Assistant 3] \{Response of Assistant 3\} [End of Assistant 3]

\State \textbf{Output Format}

\State \quad Accuracy: $\langle score_1\; score_2\; score_3\rangle$
\State \quad Reason: explanation of accuracy evaluation

\State \quad Detailedness: $\langle score_1\; score_2\; score_3\rangle$
\State \quad Reason: explanation of detailedness evaluation

\end{algorithmic}

\noindent\rule{\linewidth}{0.4pt}

\begin{figure*}
 \centering
 \captionsetup{type=figure}
 \includegraphics[width=\textwidth]
 {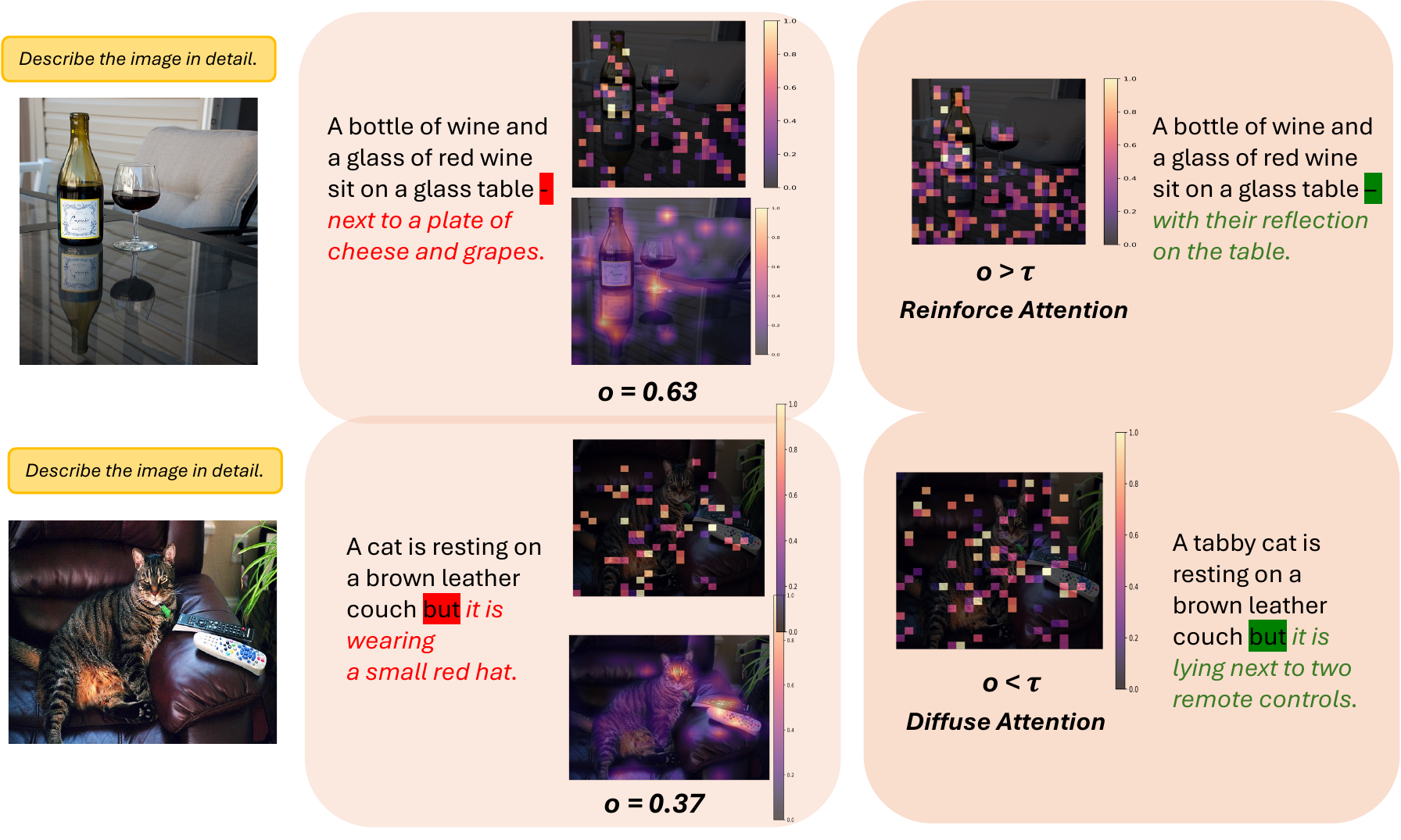}
 %\vspace{-20pt}
 \caption{Qualitative results of SAGE, including sink-token spatial attention, GradCAM-derived visual grounding maps, the computed overlap score $o$, and the modulated spatial attention maps. The sink token is highlighted in red.}
 \label{fig:supple_qual_res}
 % \vspace{-5mm}
\end{figure*}

\subsection{Runtime Analysis}
\label{sec:supp_runtime}

Since SAGE operates during inference, it introduces additional computation for grounding verification and attention modulation. The overhead arises from three components: 

\begin{itemize}
    \item  extracting key concepts from the partially generated sequence, 
    \item  computing Grad-CAM attribution from the vision encoder, and 
    \item  measuring spatial agreement between attention maps and attribution maps.
\end{itemize}
    Importantly, these operations are triggered only when sink tokens are generated rather than at every decoding step, making the additional computation sparse. All runtime measurements are conducted on a single NVIDIA A100 GPU with batch size 1.

\begin{table}
\centering
\small
\caption{Average runtime overhead introduced by different components of SAGE. Grad-CAM computation dominates the additional cost, while concept extraction and attention modulation incur negligible overhead.}
\resizebox{\textwidth}{!}{%
\begin{tabular}{lcc}
\toprule
\textbf{Component} & \textbf{Avg. Additional Cost (\%)} & \textbf{Description} \\
\midrule
Concept Extraction (POS Tagging) & 1.2 & Extract key concepts from decoded text \\
Attention Map Extraction + IoU & 1.7 & Compute spatial overlap between maps \\
Grad-CAM Attribution & 7.6 & Compute visual attribution from vision encoder \\
Attention Modulation & 0.3 & Rescale attention weights \\
\midrule
\textbf{Total SAGE Overhead} & \textbf{10.8} & Relative to standard decoding \\
\bottomrule
\end{tabular}
}
\label{tab:runtime_components}
\end{table}

    Let $T$ denote the sequence length and $k$ the number of sink tokens encountered during decoding ($k \ll T$ in practice). While standard autoregressive decoding has complexity $O(T)$, SAGE introduces an additional cost proportional to $O(k)$, keeping the overall complexity close to standard decoding. Among the added components, Grad-CAM computation accounts for most of the overhead, whereas concept extraction, spatial overlap computation, and attention modulation incur negligible cost. As observed from Table~\ref{tab:runtime_components}, we observe only an average ~10.8\% increase in total decoding time while achieving substantial reductions in hallucination.

\subsection{Additional Qualitative Results}
\label{sec:supp_qual}

\begin{figure*}
 \centering
 \captionsetup{type=figure}
 \includegraphics[width=\textwidth]
 {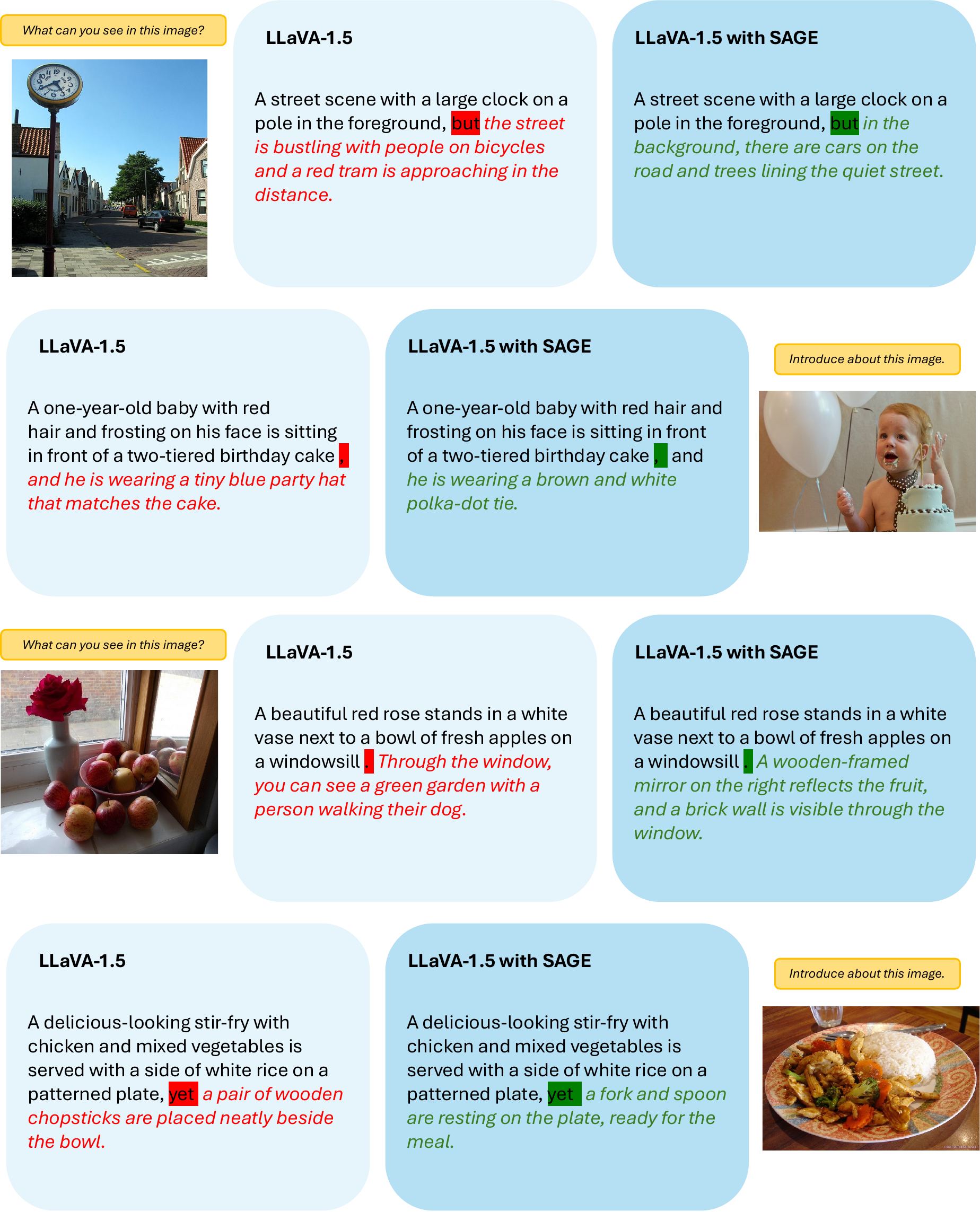}
 %\vspace{-20pt}
 \caption{SAGE's performance on mitigating hallucinations of LLaVA-1.5.}
 \label{fig:supple_qual_res_2}
 % \vspace{-5mm}
\end{figure*}

Figures~\ref{fig:supple_qual_res} and \ref{fig:supple_qual_res_2} present qualitative results illustrating the behavior of SAGE during generation. Figure~\ref{fig:supple_qual_res} visualizes spatial self-attention maps, Grad-CAM-based feature maps, and their corresponding overlap scores at sink-token checkpoints. We additionally show the modulated attention maps produced by SAGE together with the corrected generations. 

Figure~\ref{fig:supple_qual_res_2} compares the outputs of LLaVA-1.5-7B~\cite{liu2024improvedbaselinesvisualinstruction} with and without SAGE. The examples demonstrate that SAGE can effectively detect grounding inconsistencies and suppress hallucinated concepts during decoding, resulting in more visually grounded descriptions.

\subsection{Limitations}
\label{sec:supp_failure}
While SAGE effectively reduces hallucinations in generative settings such as image captioning and open-ended visual description, it has several limitations. First, the method relies on the occurrence of sink tokens during autoregressive generation to trigger grounding checks. Consequently, it is less applicable to discriminative VLM tasks (e.g., visual question answering with short answers, classification, or multiple-choice reasoning) where outputs are typically short and may not contain such structural tokens. Second, the reliability signal depends on the agreement between self-attention and gradient-based attribution; when both signals are simultaneously misaligned or noisy, the modulation mechanism may fail to correct hallucinations. Third, computing gradient-based attribution introduces additional runtime overhead, particularly for long sequences or models with large vision encoders. Finally, SAGE assumes that semantic concepts can be reliably extracted from the partially generated text, which may be challenging when the generated sequence contains ambiguous or abstract phrases. Addressing these limitations and extending grounded decoding to short-form and discriminative settings remains an important direction for future work.

\newpage

 % TODO FINAL: This \clearpage needs to be removed from both review and camera-ready versions.

% \zkn{The reviewers will ask for compute/overhead  comparisons of your method to others, so would need to provide that. }

% ZK: Note no acknowledgements due to anonymity
%\section*{Acknowledgements}
%Please insert your acknowledgments here.

% ---- Bibliography ----
%
% BibTeX users should specify bibliography style 'splncs04'.
% References will then be sorted and formatted in the correct style.
%
\bibliographystyle{splncs04}
\bibliography{main}
\end{document}